\documentclass[conference]{IEEEtran}
\IEEEoverridecommandlockouts

\usepackage{cite}
\usepackage{amsmath,amssymb,amsfonts}
\usepackage{algorithmic}
\usepackage{graphicx}
\usepackage{textcomp}
\usepackage{xcolor}
\usepackage{subcaption}
\usepackage{adjustbox}
\usepackage{hyperref}
\usepackage{cleveref}

\def\BibTeX{{\rm B\kern-.05em{\sc i\kern-.025em b}\kern-.08em
    T\kern-.1667em\lower.7ex\hbox{E}\kern-.125emX}}
\begin{document}

\bibliographystyle{IEEEtran}

\title{Towards Distribution-Shift Uncertainty Estimation for Inverse Problems with Generative Priors
\thanks{This work was supported in part by the NSF Mathematical Sciences Postdoctoral Research Fellowship under award number 2303178 to SFK. Any opinions, findings, and conclusions or recommendations
expressed in this material are those of the authors and do not necessarily reflect the views of the
National Science Foundation.}
}

\author{\IEEEauthorblockN{Namhoon Kim}
\IEEEauthorblockA{\textit{School of Electrical and Computer Engineering} \\
\textit{Georgia Institute of Technology}\\
Atlanta, Georgia 30332-0250 \\
namhoon@gatech.edu}
\and
\IEEEauthorblockN{Sara Fridovich-Keil}
\IEEEauthorblockA{\textit{School of Electrical and Computer Engineering} \\
\textit{Georgia Institute of Technology}\\
Atlanta, Georgia 30332-0250 \\
sfk@gatech.edu}
}

\maketitle

\begin{abstract}
Generative models have shown strong potential for use as data-driven priors in solving inverse problems, such as reconstructing medical images from undersampled measurements. Although these data-driven priors can improve reconstruction quality while reducing the number of required measurements, they also introduce the risk of hallucination when the image to be reconstructed falls outside the distribution of images used to train the data-driven prior. Existing approaches to uncertainty quantification in this setting (i) require an in-distribution calibration dataset, which may not be readily available, (ii) provide heuristic rather than statistical uncertainty estimates, or (iii) quantify uncertainty arising from model overparameterization or limited measurements rather than uncertainty arising from distribution shift. We propose an \emph{instance-level, calibration-set-free} uncertainty indicator that is sensitive to distribution shift, requires no prior knowledge of the training distribution, and incurs no retraining cost. 
Specifically, we posit that reconstructions of in-distribution images will be more stable with respect to variation in random measurements compared to reconstructions of out-of-distribution images, and that we can use this stability as a proxy for detecting distribution shift.
This uncertainty indicator is efficiently computable for \emph{any} inverse problem in computational imaging; we demonstrate it with preliminary experiments on tomographic reconstruction of MNIST digits, where the generative prior is a \emph{learned proximal network} trained only on digit ``0'' and evaluated on all ten digits. 
These experiments show that our uncertainty indicator, high variation among reconstructions from different measurement subsets, indeed shows higher uncertainty for out-of-distribution (OOD) digits compared to in-distribution digits. Moreover, this higher uncertainty accurately predicts the higher reconstruction error we observe for these OOD digits.
Our results motivate a deployment strategy that pairs generative priors with lightweight guardrails, to enable aggressive measurement reduction in computational imaging for in distribution images while automatically warning when the generative prior is operating out of distribution. Code is available at \url{https://github.com/voilalab/uncertainty_quantification_LPN}.
\end{abstract}


\section{Introduction}
Reconstructing images from severely undersampled measurements lies at the heart of many scientific and clinical imaging modalities. For example, in X-ray computed tomography (CT) measurements are limited because each additional projection increases carcinogenic radiation dose \cite{kalender2011computed}; in magnetic resonance imaging (MRI) each measurement is slow to collect, increasing motion blur and limiting the number of patients who can be imaged \cite{zaitsev2015motion}. Learned generative priors now offer a compelling remedy: by embedding strong, data-driven assumptions about plausible images, they can recover high-quality reconstructions from a fraction of the usual measurements--—sometimes an order of magnitude fewer—--promising faster, safer, and more accessible imaging \cite{bora2017compressedsensingusinggenerative}.

However, these benefits come with an assumption: the unseen patient must resemble the images on which the prior was trained. When that assumption fails—--because a hospital acquired a new scanner, serves a different population, or encounters a rare pathology--—the prior can \emph{hallucinate} structurally plausible but clinically erroneous details \cite{bhadra2021hallucinations}. Detecting such distribution shifts is critical. Unfortunately, today’s uncertainty quantification (UQ) toolkits offer an imperfect solution. Conformal prediction, the gold standard for statistically rigorous guarantees, requires a small calibration set drawn from the \emph{new} distribution \cite{Wen2024TaskDrivenUQ,angelopoulos2022imagetoimage,teneggi2023trust,horwitz2022conffusion,kutiel2023conformal,angelopoulos2024conformal}, which is often unavailable at deployment time. Alternatives based on bootstrap heuristics \cite{Tachella2023EquivariantBF,Wu2024UncertaintyQF,Zong2023RandomizedPM,Riis2023CUQIpyIC} or ensembling \cite{Riis2023CUQIpyP,Ramzi2020DenoisingSF,Adler2019DeepPS,Adler2018DeepBI,Jiang2022AnEA,lakshminarayanan2017simple} either sacrifice statistical validity or measure the wrong source of uncertainty (e.g. model capacity rather than distribution mismatch).

\begin{figure*}[!htbp]
  \centering
  \includegraphics[width=0.7\linewidth]{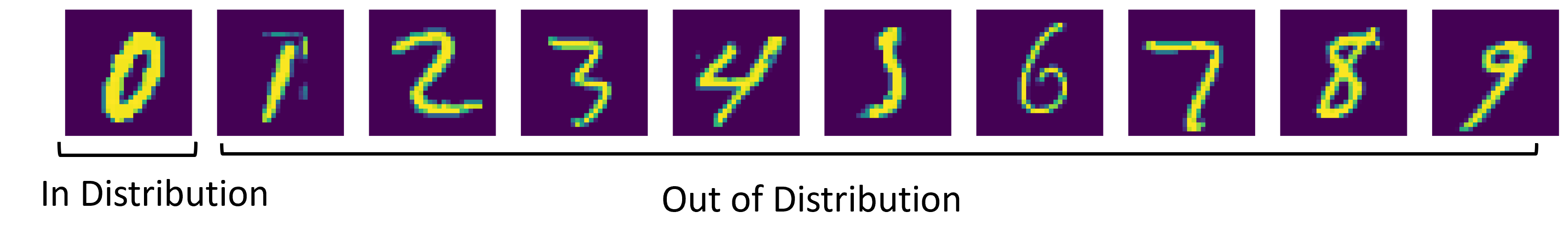}
  \caption{Our generative prior is trained on MNIST ``0'' and used for CT imaging on all digits, to test distribution shift detection.
}
\vspace{-3mm}
  \label{fig:example}
\end{figure*}

We argue that medical and computational imaging invites a simpler, calibration-set-free perspective. Each reconstruction task naturally supplies multiple measurements of the same object, for example the different projection angles measured in CT. By treating these measurements as an internal calibration set, we can build instance-level UQ signals even when no external calibration data exist. 
For a \emph{fixed set of measurements}, we randomly subsample subsets of these measurements and perform image reconstruction separately with each subset.
If the unknown imaging target is in distribution for the generative prior, we expect the reconstructions from these different subsets of measurements to be consistent with each other and with the ground truth.
However, if the unknown imaging target is out of distribution, the prior may pull the reconstruction away from the ground truth, leading to greater variations between reconstructions from different measurement subsets.
We therefore hypothesize that variation between images reconstructed from different measurement subsets can serve as a proxy to detect distribution shift.


We test this hypothesis in a toy setting of reconstructing MNIST \cite{6296535} digits from simulated sparse-view CT projections. Our generative prior is a learned proximal network (LPN) \cite{fang2024whats} trained on digit ``0'' and evaluated on all ten digits, so that digits ``1'' through ``9'' are out of distribution; see \Cref{fig:example}. We validate empirically that (i) reconstructions of in-distribution digits from different subsets of projection angles have lower variance compared to reconstructions of out of distribution digits, and (ii) this effect is most pronounced when measurements are most limited, as in this setting the reconstruction relies most on the generative prior.

These observations suggest a deployment recipe for generative priors in computational imaging that couples aggressive measurement reduction with lightweight guardrails to protect against out of distribution hallucination. The pretrained generative prior provides sample efficiency; the guardrail—--simple cross-validation between measurement subsets for each scan--—issues an automatic warning whenever the prior strays beyond its training distribution, prompting clinicians or users to acquire additional measurements or switch to a more conservative reconstruction method. 

\section{Background and Related Work}

\subsection{Sparse-View Computed Tomography (CT)}

CT is a non-invasive imaging technique that creates cross-sectional images of an object by using X-ray projections collected from multiple angles \cite{willemink2019evolution, tang2023spectral}. 
Sparse-view CT deliberately limits the number of projection angles to reduce the patient's radiation dose. However, reducing the number of projection angles leads to a severely underdetermined inverse problem, as the total number of measurements (angles \(\times\) detector resolution) becomes smaller than the number of image pixels or voxels to be reconstructed. Under these conditions, traditional reconstruction methods like Filtered Back-Projection (FBP) produce low-quality images corrupted by streak artifacts \cite{kalender2011computed}.
These artifacts can be removed by introducing a structural prior on the reconstructed image, either with an explicit geometric constraint such as low total variation \cite{candes2006robust}, or more recently by leveraging strong data-driven generative models that learn complex structure in their training images \cite{guan2023generative}.
However, these data-driven priors are sensitive to distribution shift, and can produce realistic-looking artifacts (hallucinations) on out of distribution images \cite{bhadra2021hallucinations}.

\subsection{Uncertainty and Distribution Shift in Learned Priors}
\label{sec:ood_related_work}

Uncertainty quantification methods aim to detect distribution shifts and warn users of potential hallucination from generative priors.

\paragraph{Calibration-based conformal methods}
\emph{Conformal prediction} methods leverage a small, distribution-matched calibration dataset to produce statistically rigorous finite-sample confidence intervals. In imaging, these methods span task-driven pipelines that tighten confidence intervals by acquiring more measurements \cite{Wen2024TaskDrivenUQ}, methods that produce pixelwise or masked-region confidence intervals \cite{angelopoulos2022imagetoimage, kutiel2023conformal}, diffusion-specific risk control \cite{horwitz2022conffusion, teneggi2023trust}, and distribution-shift image triage \cite{angelopoulos2024conformal}. However, these conformal prediction methods presuppose access to calibration data from the new (shifted) distribution, which limits their utility for first-encounter OOD detection.


\paragraph{Synthetic–measurement bootstraps and physics-aware methods}
Several UQ methods propose to bypass a true calibration dataset by sampling a synthetic calibration dataset from an assumed or approximate physical model.
For example, \cite{Tachella2023EquivariantBF} resample synthetic CT
projections from an initial reconstruction, and correct the
overconfidence of classical parametric bootstrap by assuming
equivariance across the CT nullspace.  
For physical systems governed by partial differential equations (PDEs), several methods propose deterministic physics
surrogates together with latent-space uncertainty evolution, \cite{Wu2024UncertaintyQF,Zong2023RandomizedPM,Riis2023CUQIpyIC}
While these techniques do not require a distributionally matched calibration dataset, they inherit any bias or errors in
the physics surrogate or the synthetic model, and lack rigorous guarantees of validity for the resulting confidence intervals.

\paragraph{Bayesian and ensemble methods}
Sampling from an explicit or implicit posterior distribution---using Markov Chain Monte Carlo (MCMC) sampling~\cite{Riis2023CUQIpyP,Ramzi2020DenoisingSF} or its stochastic gradient variant (SG-MCMC)~\cite{Adler2019DeepPS,Adler2018DeepBI}  over a
learned prior---captures \emph{epistemic} uncertainty due to limited measurements but ignores uncertainty due to distribution mismatch.
Ensemble‐based approximations~\cite{Jiang2022AnEA, lakshminarayanan2017simple} can capture uncertainty over trained model weights,
but all members of the ensemble share the same training data and can therefore fail in unison when reconstructing OOD images.

\subsection{Learned Proximal Networks as Generative Priors}
\label{sec:lpn}


During image reconstruction, a regularizer $R$ is often enforced through a proximal operator $f=\operatorname{prox}_{\lambda R}$, which moves the current iterate towards the prior after each step of optimization. Every proximal operator is the gradient of a convex function, so Learned Proximal Networks (LPNs) \cite{fang2024whats} learn an input-convex neural network $\psi_\theta$ and apply its gradient $f_{\theta}(z)=\nabla_{z}\psi_{\theta}(z)$ as the proximal operator of the corresponding implicit learned regularizer $R_\theta$.
Using $x^{(k)} \in \mathbb{R}^n$ to denote the $k^{\text{th}}$ iterate of a proximal method, $v^{(k)}$ as intermediate iterates, $A \in \mathbb{R}^{m\times n}$ as the measurement model (e.g. the Radon transform for CT), and $y = A x^{\star} \in \mathbb{R}^m$ as the measurements (e.g. projections for CT), we can write an iteration of a proximal gradient algorithm as
\begin{equation}\label{alg:proximal}
v^{(k)} = x^{(k)} - \eta_k A^{\!\top}(Ax^{(k)}-y),
\quad
x^{(k+1)} = f_{\theta}\!\bigl(v^{(k)}\bigr),
\end{equation}
where $\eta_k$ is a step size and the data-driven prior is injected through the proximal operator $f_{\theta}$.

To learn $f_{\theta}$ from a training dataset, \cite{fang2024whats} proposes the
\emph{proximal matching loss} for samples from unknown distribution $p_{x}$. Given noised example ${z} = {x} + \sigma {\varepsilon}$; ${x} \sim p_{x}$, ${\varepsilon} \sim\mathcal{N} (0,{I_n)}$, 
 \begin{equation}\label{eq:prox-match}
\mathcal L_{\mathrm{PM}}(\theta;\gamma)
= \mathbb E_{\mathbf{x,y}}\!\left[m_{\gamma}\!\left(\|f_{\theta}({z})-{x}\|_{2}\right)\right],
\end{equation}
\vspace{-4mm}
\begin{equation*}
\text{where}~~m_{\gamma}(r)
= 1-\frac{1}{(\pi\gamma^{2})^{n/2}}
   \exp\!\left(-\frac{r^{2}}{\gamma^{2}}\right)
\end{equation*}
and $\gamma>0$ controls how sharply $m_{\gamma}$ approximates a
Dirac delta.  Minimizing~\eqref{eq:prox-match} over the training dataset maximizes the posterior density $p_{{x}|{z}}(f_{\theta}({z}))$. As $\gamma\!\to\!0$, $f_{\theta}$ converges
to the maximum a posteriori (MAP) denoiser.


\section{Methods}
\label{sec:methods}

Let $y = A x^{\star} + \epsilon$ be the noisy sinogram acquired by a
fan-beam CT scanner with a $22$-pixel detector.
Here $A\in\mathbb R^{m\times n}$ is the forward operator (Radon transform),
$x^{\star}\in\mathbb R^{n}$ is the unknown image or X-ray attenuation map
($n\!\gg\!m$ in sparse-view CT), and
$\epsilon$ represents noise. In our experiments, $x^\star \in [0,1]^{28\times 28}$ is an MNIST digit  and we draw $\epsilon$ from an isotropic mean-zero Gaussian distribution with standard deviation $\sigma\!=\!2$.
We train an LPN on digit ``0'' images, and use the same trained LPN throughout all experiments.
We consider three measurement budgets to constrain the number of CT projections: $N_{\text{view}}\in\{11,22,33\}$; all of these regimes are undersampled, with fewer total measurements compared to the number of pixels in the target image.
For each measurement budget $N_{\text{view}}$, we repeat the measurement process 10 times with different random seeds. This construction 
allows us to evaluate our proposed OOD metric by computing pixel-wise variance across images reconstructed from different sets of random measurements.

Once the LPN is trained on a training dataset of digit ``0'', we simulate random CT measurements $y$ for each of ten randomly selected MNIST images from each of the ten digits ($100$ images total, all unseen during LPN training) for evaluation. For each set of random measurements, we reconstruct an image following a proximal algorithm (\Cref{alg:proximal}) with our LPN as the proximal operator. 
We evaluate reconstruction quality using PSNR and SSIM \cite{1284395} compared to the true MNIST images, and evaluate our proposed metric of pixel-wise variance between reconstructions of the same image from different random measurements.

\section{Experiments}
\label{sec:experiments}

\begin{figure}[!h]
  \centering
  \includegraphics[width=\linewidth]{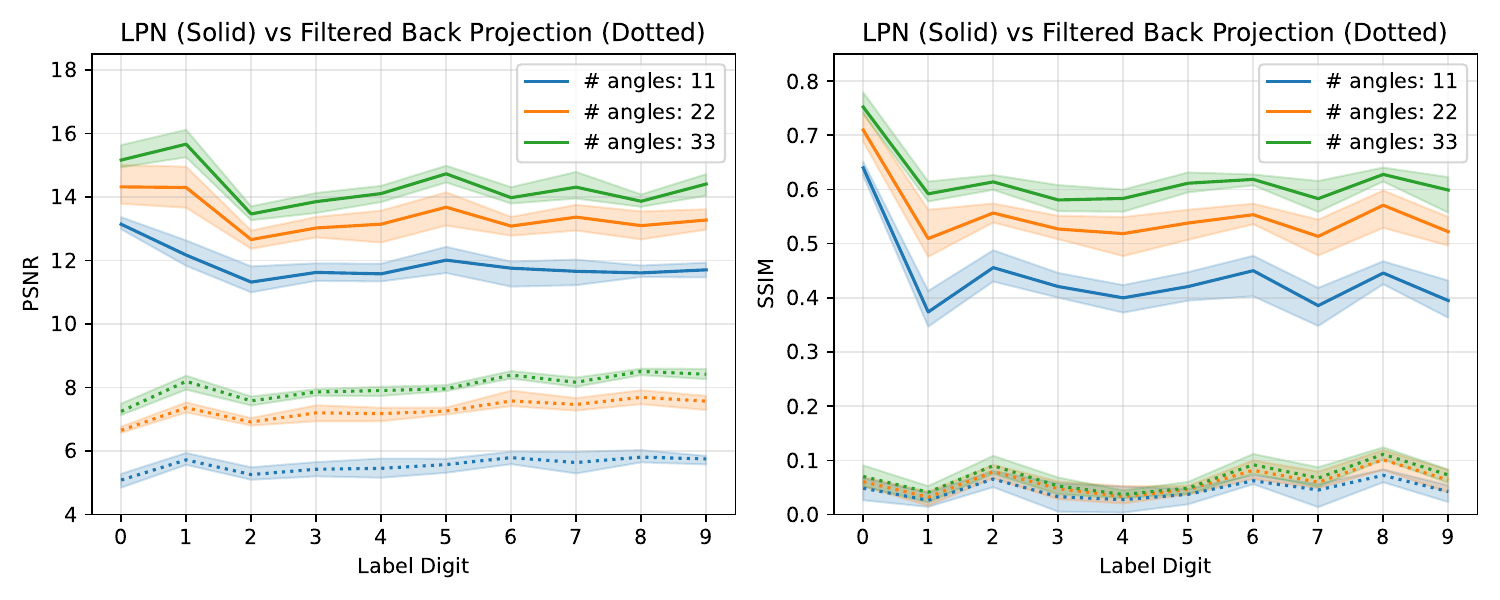}
  \caption{Per-digit mean PSNR and SSIM (averaged over 10 images with 10 random seeds per image) with error bars indicating the min–-max spread. Solid curves correspond to LPN reconstructions and dashed curves to the FBP baseline; LPN consistently outperforms FBP. The performance gap is largest for the in-distribution digit ``0'', whose PSNR/SSIM is higher than those of OOD digits when reconstruction leverages the LPN. This is especially pronounced in the 11-view experiment that is most undersampled and thus relies most heavily on the learned prior.}

  \label{fig:ct}
\end{figure}

\begin{figure*}[tbp]
  \centering
  \begin{subfigure}{0.9\linewidth}
    \centering
    \begin{adjustbox}{center}
    \includegraphics[width=\linewidth]{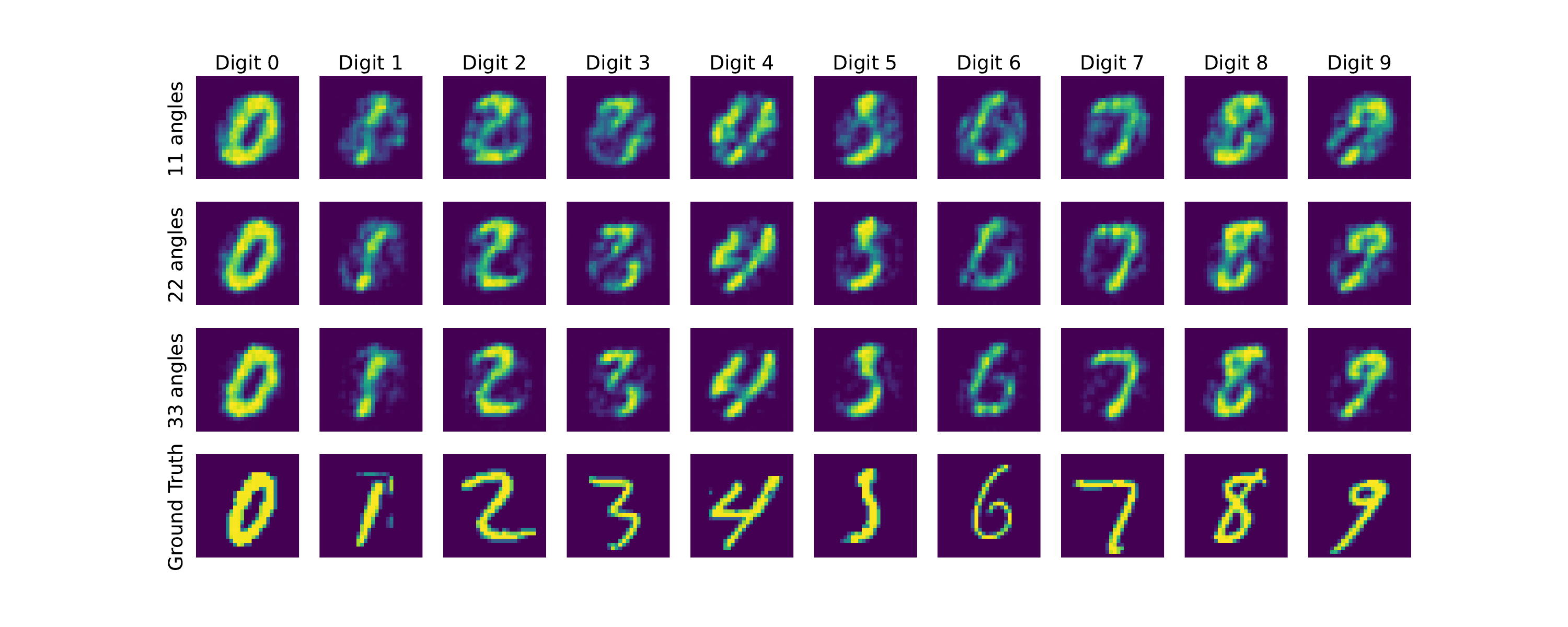}
  \end{adjustbox}
    \vspace{-2em} 
    \caption{Grid of “mean” images for one example of each digit: for each digit and budget of projection angles, we plot the pixel-wise average of the reconstructions over the 10 seeds. These average reconstructions show greater consistency across random seeds for the in-distribution digit ``0'', especially when the number of measurement angles is severely limited.
    }
    \label{fig:mean}
  \end{subfigure}

  \vspace{-0em} 

  \begin{subfigure}{0.9\linewidth}
    \centering
    \begin{adjustbox}{center}
    \includegraphics[width=\linewidth]{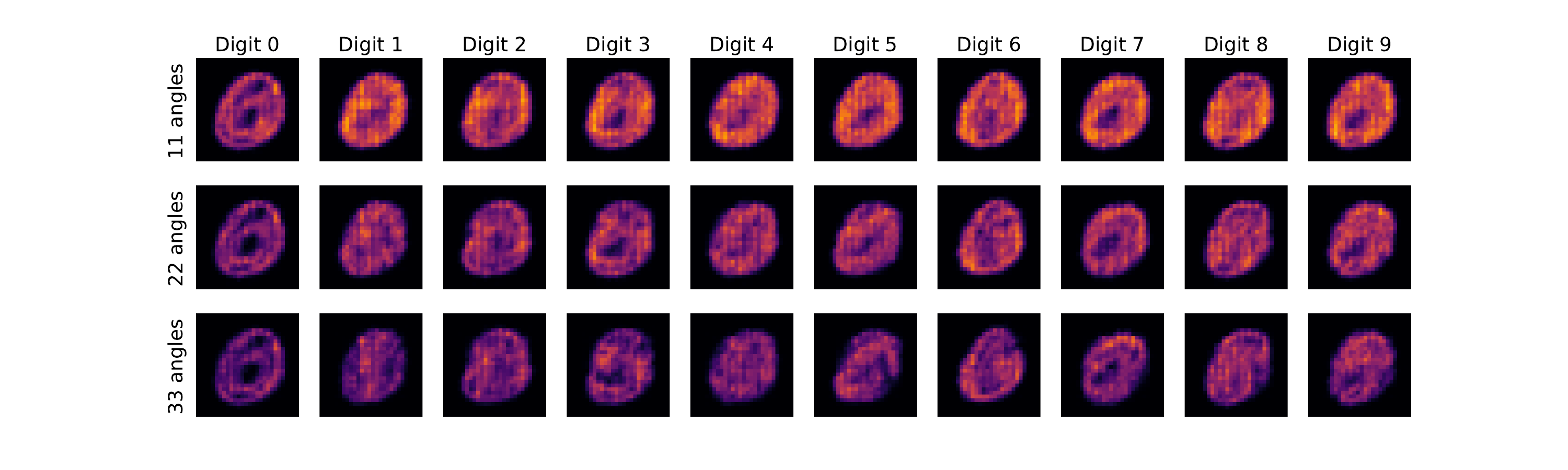}
  \end{adjustbox}
  \vspace{-2em} 
    \caption{Heat maps of pixel-wise standard deviation across the same 10 seeds, highlighting that in distribution reconstructions stay consistent across different sets of random measurements while OOD digits show large variability in reconstructions, especially with only 11 projection measurements. This higher pixel-level variance for OOD reconstructions validates our hypothesis and serves as an indicator to detect distribution shift on a single image.}
    \label{fig:std}
  \end{subfigure}
    
  \caption{Visualizing distribution shift detection on MNIST: (a) mean reconstructions, (b) pixel-wise standard deviation.}
  \label{fig:grouped_mean_std}
\end{figure*}

\begin{figure}[h]
  \centering
  \includegraphics[width=0.7\linewidth]{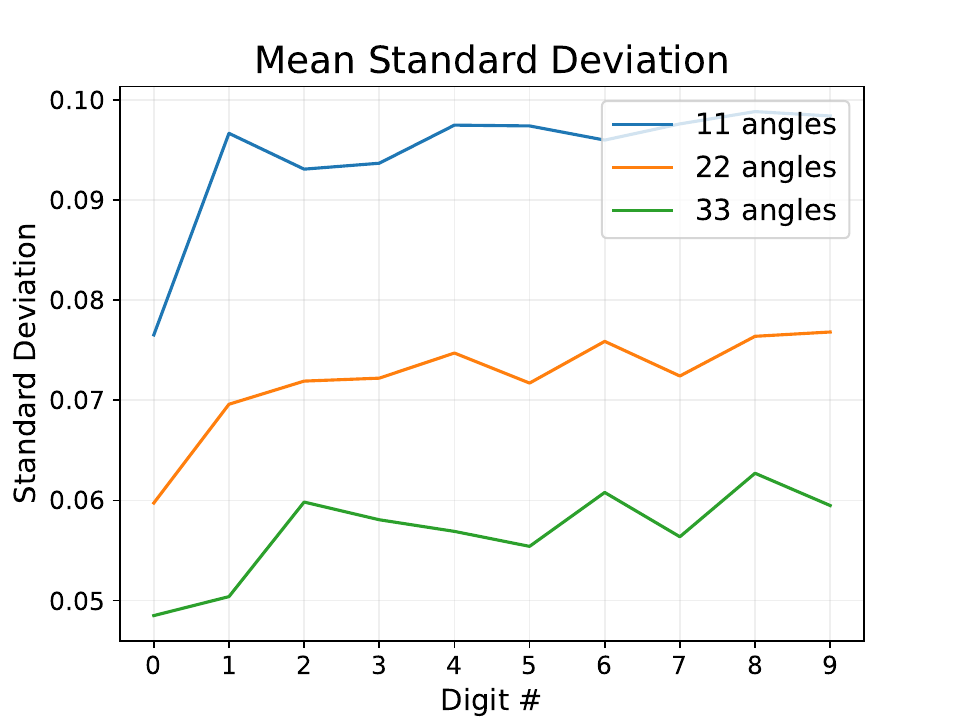}
  \caption{Average pixel-wise standard deviation is lower for the in-distribution digit ``0'' than the OOD digits, confirming our hypothesis that reconstruction instability across random measurements can detect distribution shift.
}
  \label{fig:std_plot}
  \vspace{-2mm}
\end{figure}



We begin by comparing the reconstruction quality an LPN achieves for sparse-view CT of in-distribution versus OOD images.
In \Cref{fig:ct} we plot the mean and range (min--max) PSNR and SSIM of the 100 reconstructions for each digit, including 10 random measurement seeds for each of the 10 MNIST images selected for each digit.
We compare reconstructions from the proximal method with our learned prior (trained on digit ``0'') against a standard unregularized FBP baseline.
The results indicate that the learned prior is beneficial even for out of distribution digits, but much more effective for in-distribution digits, and especially so as the number of measurement angles is reduced.

We evaluate our proposed distribution shift metric, variation across reconstructions from different measurements, qualitatively in \Cref{fig:grouped_mean_std} and quantitatively in \Cref{fig:std_plot}.
If the target image is in the distribution learned by the prior network, we expect it to produce consistent predictions even as the set of random measurement angles changes. In contrast, if the target image is out of distribution for the prior, we expect higher variance of the reconstructions when the random measurement angles change. This is exactly what we find: \emph{reconstruction variance over random measurements detects distribution shift}. The effect is most prominent when the number of measurement angles is small, which aligns with the setting when the learned prior has the most influence on the reconstruction and thus distribution shift poses the greatest risk.




\section{Discussion}

Generative models have shown great promise as data-driven priors in solving inverse problems like CT reconstruction, enhancing image quality and reducing measurements. However, data-driven priors pose risks of hallucination under distribution shift, when the target image differs from the distribution used to train the prior. Here we validate the simple hypothesis that this distribution shift fragility can be detected without extensive computational or data-collection burden, by evaluating how consistent the reconstruction is across random subsets of the available measurements. Our work suggests a simple strategy to detect and mitigate distribution shift by collecting additional measurements until reconstruction stability crosses a desired threshold.

\paragraph{Limitations and future work} Though our proposed distribution shift uncertainty estimator is broadly applicable across inverse problems, our initial experimental validation is preliminary and limited to the toy setting of tomographic reconstruction of MNIST digits using a learned proximal network as data-driven prior. It is a high priority for future work to evaluate the potential of our proposed uncertainty metric on diverse datasets of practical significance in different imaging inverse problems, and with diverse data-driven priors including diffusion models. 
We also acknowledge that there may be settings in which our proposed uncertainty indicator may not correlate accurately with distribution shift. For example, an otherwise in-distribution image that is noisier than the training dataset may be erroneously flagged as OOD by our metric. Conversely, if the portion of an image that is OOD does not contribute to the measurements (i.e. if the distribution shift is correlated with the forward model), our metric would have no way to detect it as OOD. Addressing cases like these is also a high priority for future work.
Finally, we encourage future work to consider statistical analysis of our proposed OOD uncertainty metric, to build valid and robust confidence intervals and offer guidance on how many (sets of) measurements to employ for the most accurate OOD detection.

\newpage
\bibliography{IEEEabrv,references}

\end{document}